\documentclass[twoside,11pt]{article}

\usepackage{jmlr2e}

\usepackage{float}
\usepackage{lscape}
\usepackage{multirow, pbox}
\usepackage{algorithm}
\usepackage{algpseudocode}
\usepackage{amsmath}
\usepackage{amssymb}

\usepackage{caption}
\usepackage{subcaption}
\usepackage{fullpage}

\usepackage{amsfonts}

\usepackage{caption}
\usepackage[utf8]{inputenc}
\usepackage{longtable}

\usepackage{lscape}
\usepackage{pdflscape}
\usepackage{multirow}

\firstpageno{1}

\begin{document}

\title{Radius-margin bounds for deep neural networks}

\author{\name Mayank Sharma \email eez142368@iitd.ac.in \\
       \addr Department of Electrical Engineering\\
       \addr Indian Institute of Technology, Delhi\\
       \addr New Delhi, 110016, India
       \AND
       \name Jayadeva \email jayadeva@ee.iitd.ac.in \\
       \addr Department of Electrical Engineering\\
	   \addr Indian Institute of Technology, Delhi\\
	   \addr New Delhi, 110016, India
	   \AND
	   \name Sumit Soman \email eez127509@ee.iitd.ac.in \\
	   \addr Department of Electrical Engineering\\
	   \addr Indian Institute of Technology, Delhi\\
	   \addr New Delhi, 110016, India}

\editor{}

\maketitle

\begin{abstract}
Explaining the unreasonable effectiveness of deep learning has eluded researchers around the globe. Various authors have described multiple metrics to evaluate the capacity of deep architectures. In this paper, we allude to the radius margin bounds described for a support vector machine (SVM) with hinge loss, apply the same to the deep feed-forward architectures and derive the Vapnik-Chervonenkis (VC) bounds which are different from the earlier bounds proposed in terms of number of weights of the network. In doing so, we also relate the effectiveness of techniques like Dropout and Dropconnect in bringing down the capacity of the network. Finally, we describe the effect of maximizing the input as well as the output margin to achieve an input noise-robust deep architecture.
\end{abstract}


\section{Introduction}
Deep neural networks (DNN) have been the method of choice for owing to their great success in plethora of machine learning tasks, such as image classification and segmentation \citep{krizhevsky2012imagenet}, speech recognition \citep{hinton2012deep}, natural language processing \citep{collobert2011natural}, reinforcement learning \citep{mnih2015human} and various other tasks \citep{schmidhuber2015deep,lecun2015deep}. It is known that depth and width of network plays a key role in its learning abilities. Although multiple architectures of DNNs exits like recurrent neural networks (RNNs) \citep{hochreiter1997long} and recursive nets \citep{socher2011parsing} however, for the discussions in the paper, we focus on the feed-forward architectures of DNNs. In the works of \citep{hornik1991approximation,cybenko1989approximation} it was shown that a single hidden layer network or a shallow architecture can approximate any measurable Borel function given enough number of neurons in the hidden layer but recently it was shown by \citep{montufar2014number} that a deep network can divide the space into an exponential number of sets, a feat which cannot be achieved by a shallow architecture with same number of parameters. Similarly, it was shown by \citep{telgarsky2016benefits} that for a given depth and the number of parameters there exists a DNN that can be approximated by a shallow network with parameters that are exponential in number of layers of the network. \citep{cohen2016expressive} conclude that functions that can be implemented by DNNs are exponentially more expressive than functions implemented by a shallow network. These theoretical results which showcase the expressiveness of DNNs have been empirically backed up with deep architectures being the current state-of-the-art in multiple applications across various domains. 

Many researchers have shown the effect of depth and width on the performance of deep architectures. It is known that increasing the depth or width increases the number of parameters in the network, and often these numbers can be much larger than the number of the samples used to the train the network itself. These networks are currently trained using stochastic gradient descent (SGD). With such a huge number, the obvious question is to ask is why do these machines learn effectively? Researchers have tried to answer this question by proving statistical guarantees on learning capacities of these networks. Multiple complexity measures have been proposed in the literature, namely Vapnik-Chervonenkis (VC) dimension \citep{vapnik1998statistical} and its related extensions like pseudo-dimension, fat shattering dimension \citep{anthony2009neural} and radius-margin bounds \citep{burges1998tutorial,smola1998learning}, Radamacher Complexity \citep{bartlett2002rademacher} and covering numbers \citep{zhou2002covering} to name a few. All these measures define a number that characterizes the complexity of the hypothesis class which in this case is the neural network. The most popular among these is the VC dimension which defines the size of largest set that can be \textit{shattered} by the given hypothesis class.

\citep{bartlett1999almost} provided the VC dimension bounds for piecewise linear neural networks. \citep{karpinski1995polynomial,sontag1998vc,baum1989size} gave the VC bounds for general feed-forward neural network with sigmoidal non-linear units. These bounds are defined with respect to the number of parameters and in general are quite large. \cite{shalev2014understanding} presented bounds which are linear in terms of the trainable parameters. These bounds grow larger with increase in width and depth of the networks and fail to explain the unreasonable effectiveness of the depth of neural networks. \citep{bartlett1998sample} showed that the VC dimension of a network can be bounded by the norm of parameters or weights rather than the number of parameters. The norm of weights can be much smaller than the number of weights. Thus this bound could explain the rationale behind the minimization of the norm of the weights. \citep{neyshabur2015norm} presented Radamacher complexity based bounds showing the deep network bounds in terms of norm of the weights and the number of neurons per layer. \citep{sun2016depth} also presented Radamacher average bounds for multi-class convolutional neural networks with pooling operations in terms of norm of the weights and the size of pooling regions. \citep{xie2015generalization} showed that mutual angular regularizer (MAR) can greatly improve the performance of a neural network. They showed that increasing the diversity of hidden units in a neural network reduces the estimation error and increases the approximation error. Authors also presented generalization bounds in terms of Radamacher complexity. However, as mentioned in \citep{kawaguchi2017generalization}, the dependency on the depth of the network is exponential. \citep{sokolic2017robust} presented generalization bounds in terms of Jacobian matrix of the network and showed better performance of networks when presented with smaller number of samples. They provide theoretical justification to the contractive penalty used in \citep{an2015contractive,rifai2011contractive} by explaining the effect of Jacobian regularization on input margin.

Currently, the neural networks are regularized using Dropout \citep{srivastava2014dropout} or Dropconnect \citep{wan2013regularization} in conjugation with weight regularization. Dropout randomly drops the neurons to prevent their co-adaptations, while Dropconnect randomly drops connections trying to achieve a similar objective. Both these methods can be thought of as ensemble averaging of multiple neural networks done through a simple technique of using Bernoulli gating random variables to remove certain neurons and weights respectively. The properties of Dropout are studied in \citep{baldi2014dropout} while the Radamacher complexity analysis of dropconnect is mentioned in \citep{wan2013regularization}.

There have also been performance analysis of various architectures of feed-forward neural networks. One such architecture is residual network (resnet) \citep{he2016deep}, whose analysis has been presented in \citep{he2016identity}. It uses a direct or an identity connection from previous layer to the next layer and allows very deep architectures to be trained effectively with minimal gradient vanishing problem \citep{bengio1994learning}. Several variants of residual networks have been proposed namely wide residual networks \citep{zagoruyko2016wide}, inception residual network \citep{szegedy2017inception} and the generalization of residual networks named as highway networks \citep{srivastava2015highway}.

Our contribution in this work is as follows, Firstly we present radius-margin bounds on feed-forward neural networks. Then, we show the bounds for Dropout and Dropconnect and show that these regularizers brings down the expected sample complexity for deep architectures. Next, we present the margin bounds for residual architectures. Furthermore, we compute the radius-margin bound for an input noise-robust algorithm and then show that Jacobian regularizer along with hinge loss approximates the input noise-robust hinge loss. Finally, we hint at the fact that enlarging the input margin for a neural network via minimization of Jacobian regularizer is required to obtain an input noise-robust loss function. To our knowledge this is one of the first effort in showing the effectiveness of various regularizers in bringing down the sample complexity of neural networks using the radius margin bounds for margin based loss function. In this paper make use of binary class support vector machine (SVM) \citep{cortes1995support} loss function at the output layer. This can also be generalized to any margin based loss function for both binary and multi-class settings.

\section{Preliminaries}
\label{sec:prelim}
Given a binary class problem, we denote $\mathcal{X} \in \Re^{d}$ as the input set and $\mathcal{Y} \in \{-1,1\}$ as our label set. Here $d$ is the dimensionality of the input pattern. The training set is defined as $S=((\mathbf{x}^1,y_1),\ldots,(\mathbf{x}^m,y_m))$, which is a finite sequence of $m$ pairs in $\mathcal{X} \times \mathcal{Y}$. Let $\mathcal{D}$ denote the probability distribution over $\mathcal{X} \times \mathcal{Y}$. The training set $S$ is i.i.d. sampled from the distribution $\mathcal{D}$. Let $\mathcal{F}$ denotes the hypothesis class. The goal of the learning problem or a learning algorithm $\mathcal{A}$ is to find a function $f \in \mathcal{F} : \mathcal{X} \times \mathcal{Y} \rightarrow \Re$. Let $\gamma$ denote the margin of classifier, which is given by:
\begin{gather}
\gamma = \sup_a \min_i \inf_\mathbf{x} \{ \|\mathbf{x}^i-\mathbf{x}\|_2 \leq a : f(\mathbf{x})=0;\; y_if(\mathbf{x}^i) > 0 \}  \nonumber \label{eq:margin} 
\end{gather}
Now, consider a feed-forward neural network with one input layer, $P$ hidden layers and one output neuron. Let the number of neurons in layer $k$ be given by $h_k : k \in \{0,\ldots,P+1\}$, where $h_0 = d$ denote the dimension of the input sample and $h_{P+1} = 1$ denote the number of neurons in output layer. The number of units in output layer is one for binary classification. Let $\mathbf{W}_{\{k,k+1\}} \in \Re^{\{h_k,h_{k+1}\}}$ denote the weights going from layer $k$ to layer $k+1$ such that, $\mathbf{w}_{\{k,k+1\}}^t \in \Re^{h_k}$ denote the weights going from layer $k$ to neuron $t \in \{1,\ldots,h_j\}$ in layer $k$. Let $\sigma(\cdot)$ denote the activation function, which is Rectified Linear Units (ReLUs), tanh or any $L_\sigma$ Lipschitz continuous activation function passing though origin for each of the neurons in hidden layers and a linear activation function in output layer. We keep the norm of inputs bounded by $R$ i.e., $\|\mathbf{x}\|_2 \leq R,\; \forall\; x \in \mathcal{X}$ and the norm of weights going from layer $k$ to layer $k+1$ bounded by $A_k$ such that $\|\mathbf{w}_{\{k,k+1\}}^t\|_2 \leq A_k$. The function computed by network in layer $k$ is given by $\phi(\mathbf{x})_k = \sigma(\phi(\mathbf{x})_{k-1} \cdot W_{\{k-1,k\}}) \in \Re^{h_k},\; \forall\; k \in \{1,\ldots,P\}$, where $\phi(\mathbf{x})_0 = \mathbf{x} $, $\phi(\mathbf{x})_{P+1} = \phi(x)_{P} \cdot \mathbf{w}_{\{P,P+1\}}^1 $, $\sigma(\cdot)$ is applied element-wise and the operator $\cdot$ denotes the dot product.

Thus, the hypothesis class of feed-forward neural networks with $P$ hidden layers and norm of weights bounded by $A_j$ is given by:
\begin{gather}
\mathcal{F} = \sigma(\sigma(\ldots \sigma(\mathbf{x} \cdot \mathbf{W}_{\{0,1\}}) \cdot \mathbf{W}_{\{1,2\}} \ldots \cdot \mathbf{W}_{\{P-2,P-1\}}) \cdot \mathbf{W}_{\{P-1,P\}})\cdot \mathbf{w}_{\{P,P+1\}}^1 \nonumber\label{eq:F}
\end{gather}
\textbf{Lipschitz property}: Let $C \subseteq \Re^d$. A function $\sigma : \Re^d \rightarrow \Re^k$ is $L_\sigma$-Lipschitz over $C$ if for every $\mathbf{x}^1, \mathbf{x}^2 \in C$, we have:
\begin{gather}
\|\sigma(\mathbf{x}^1)-\sigma(\mathbf{x}^2)\|_2 \leq L_\sigma\|\mathbf{x}^1-\mathbf{x}^2\|_2 \label{eq:lipschitz}
\end{gather}
ReLU $\sigma_{ReLU}(z) = max(0,z)$ and Leaky ReLU $\sigma_{LReLU}(z) = max(0.01z,z)$ are Lipschitz continuous function with Lipschitz constant $L_\sigma = 1$. Likewise, sigmoid $\sigma_{sig}(z) = \frac{1}{1+\exp(-z)}$ and hyperbolic tangent $\sigma_{tanh}(z) = \frac{\exp(x) - \exp(-x)}{\exp(x) + \exp(-x)}$ are Lipschitz continuous functions with Lipschitz constants $L_\sigma = \frac{1}{4}$ and $L_\sigma = 1$ respectively. We will focus mostly on ReLU and activation functions which passes through origin like Leaky ReLU and tanh.

For activation functions passing through origin, eq. \ref{eq:lipschitz} holds true for all $z^1, z^2 \in C$. Hence, the eq. \ref{eq:lipschitz} also holds for all $z^1 \in C$ and $z^2 = 0$
\begin{gather}
\|\sigma(z^1)-\sigma(z^2)\|_2 \leq L_\sigma\|z^1-z^2\|_2 \nonumber \\ 
\|\sigma(z^1)-\sigma(0)\|_2 \leq L_\sigma\|z^1-0\|_2 \nonumber\\
\text{since} \; \sigma(0) = 0 \; \text{for functions passing through origin} \nonumber \\
\|\sigma(z^1)\|_2 \leq L_\sigma\|z^1\|_2  \label{eq:lipschitz_norm}
\end{gather}
\textbf{$0-1$ loss}: The $0-1$ loss for random variable $(\mathbf{x},y) \in \mathcal{X} \times \mathcal{Y}$ and predictor $ g(\mathbf{x}) = sign(f(\mathbf{x})) : f \in \mathcal{F}$ is given by:
\begin{gather}
\ell_{0-1}(g,(\mathbf{x},y)) = \begin{cases}
0 \;\; \text{if} \; g(\mathbf{x}) = y\\
1 \;\; \text{if} \; g(\mathbf{x}) \neq y
\end{cases} \label{eq:l 0-1}\nonumber
\end{gather}
\textbf{True risk of $0-1$ loss}: The true risk of the prediction rule $g$ is defined as:
\begin{gather}
L_\mathcal{D}^{0-1}(g) = \mathbb{P}_{(\mathbf{x},y) \sim \mathcal{D}}[g(\mathbf{x}) \neq y] \label{eq:LD 0-1} \nonumber
\end{gather}
\textbf{Empirical risk of $0-1$ loss}:  The empirical risk of the prediction rule $g$ is defined as:
\begin{gather}
L_S^{0-1}(g)= \frac{1}{m}\sum_{i=1}^{m}(g(\mathbf{x}^i) \neq y_i) \label{eq:LS 0-1} \nonumber
\end{gather}
\textbf{Hinge loss}: The empirical $0-1$ risk is difficult to optimize owing to its non-convex nature. Hinge loss satisfies the requirements of a convex surrogate for $0-1$ loss. The hinge loss is defined as:
\begin{gather}
\ell_{hinge}(f,(\mathbf{x},y)) = \max(0, 1 - yf(\mathbf{x})) \label{eq:l hinge} \nonumber
\end{gather}
Clearly, $\ell_{0-1}(g,(\mathbf{x},y)) \leq \ell_{hinge}(f,(\mathbf{x},y))$.\\
\textbf{Empirical risk of hinge loss}:  The empirical risk of $\ell_{hinge}(f,(x,y))$ is defined as:
\begin{gather}
L_S^{hinge}(f)= \frac{1}{m}\sum_{i=1}^{m}\max(0, 1 - y_if(\mathbf{x}^i)) \label{eq:LS hinge} \nonumber
\end{gather}

\section{Radius-margin bounds}
\label{sec:radius_margin}
In this section we provide radius margin bounds for feed-forward neural networks including those of regularizers like Dropout and Dropconnect. The reader should refer to the section \ref{Appendix} (Appendix) for the proofs of the theorems mentioned in the main text.
\begin{theorem} \label{th:1}
	The upper bound on VC dimension $VCdim(\mathcal{F})$ for a training set $S \subseteq \{(\mathbf{x},y)^m : \|x\|_2 \leq R;\; (\mathcal{X} \times \mathcal{Y})^m\}$ which is fully \textit{shattered} with the output margin of $\gamma = \frac{1}{\| \mathbf{w}_{\{P,P+1\}}^1\|_2}$ by a function $f \in \mathcal{F}$ from the hypothesis class $\mathcal{F}$ of neural networks with $P$ hidden layers, $h_k,\; \forall\; k \in \{0,\ldots,P+1\}$ neurons in each layer with $L_\sigma$-Lipschitz activation function passing through origin and the norm of weights constrained by $A_k$ for all $k \in \{1,\ldots,P+1\}$ is given by:
	\begin{gather}
	VCdim(\mathcal{F}) \leq R^2A_{P+1}^2 L_\sigma^{2P}\prod_{k=1}^{P}h_kA_k^2 \label{eq:VC FNN1}
	\end{gather}
\end{theorem}

The bound given in eq. \ref{eq:VC FNN1} defines the dependence of VC dimension on the radius of data and the product of max-norm terms with number of neurons per layer. There is always a dependence on the depth of the network in terms of number of product terms included in the eq. \ref{eq:VC FNN1}. The bound has several implications:
\begin{enumerate}
	\item The bound is independent of dimensionality on input data, it is only dependent on radius of data.
	\item The bound is independent of number of weights, but rather dependent on max-norm of weights.
	\item The bound depicts the key role of depth for deep networks. Increasing the depth of the network does not always increase the VC dimension. If the product $h_kA_k^2 \leq 1$ for all $k = \{1,\ldots,P\}$ then there is a decrease in the capacity of the network as the depth increases. On the other hand if the product $h_kA_k^2 \geq 1$ for all $k = \{1,\ldots,P\}$, then the network capacity increases with depth. Thus, by changing the number of neurons and max-norm constraints on weights one can alter the capacity of the network to the desired values. 
	\item Keeping the number of neurons in hidden layers fixed to $h$ and using ReLU activation function with $L_\sigma = 1$, we get the VC bound similar to Theorem 1 of \citep{neyshabur2015norm}:
	\begin{gather}
	VCdim(\mathcal{F}) \leq R^2A_{P+1}^2 h^P \prod_{k=1}^{P}A_k^2 \label{eq:VC FNN2}
	\end{gather}
	The bound presented in eq. \ref{eq:VC FNN2} shows that keeping the number of neurons fixed in each layer, the VC dimension of the hypothesis class of neural network can be controlled by changing the max-norm constraint on the weights on the network. However, the exponential dependency on the depth cannot be avoided.
\end{enumerate}
\textbf{Effect of Dropout}: We now show the effect of Dropout on the same network, where we multiply each neuron $n$ in layer $k$ with Bernoulli selector random variable $\delta_{\{k,n\}}^i$ for each sample $i$. Every selector random variable $\delta_{\{k,n\}}^i$ takes the value $1$ with probability $p_k$ and $0$ with dropout probability $q_j = 1 - p_j$ for each layer $j$ and is independent from each other. The dropout mask for each layer $j$ and each sample $i$ is given by $\mathbf{u}_k^i \in \{0,1\}^{h_k}$, where each entry of the mask is $u_{\{k,n\}}^i = \delta_{\{k,n\}}^i$. The new hypothesis class $\mathcal{F}_{do}$ of neural network is given as:
\begin{gather}
\begin{split}
\mathcal{F}_{do} =   (\mathbf{u}_{P} \odot \sigma( (\mathbf{u}_{P-1} \odot \sigma(\ldots (\mathbf{u}_1 \odot \sigma( (\mathbf{u}_0 \odot \mathbf{x}) \cdot \mathbf{W}_{\{0,1\}}) \\ \cdot \mathbf{W}_{\{1,2\}}) \ldots \cdot \mathbf{W}_{\{P-2,P-1\}})) \cdot \mathbf{W}_{\{P-1,P\}}))\cdot \mathbf{w}_{\{P,P+1\}}^1
\end{split} \nonumber
 \label{eq:F-dropout}
\end{gather}
Here, $\odot$ represents element wise multiplication.
\begin{theorem}\label{th:2}
	 For the same network as mentioned in Theorem \ref{th:1}, when added with dropout to each layer $k$ for all  $k \in \{1,\ldots,P+1\}$ with dropout probability $q_k$, the VC dimension bound $VCdim(\mathcal{F}_{do})$ is bounded by:
	 \begin{gather}
	 	VCdim(\mathcal{F}_{do}) \leq p_{0}R^2A_{P+1}^2 L_\sigma^{2P}\prod_{k=1}^{P}p_kh_kA_k^2 \label{eq:VC FNN_DO1}
	 \end{gather}
\end{theorem}
\textbf{Effect of Dropconnect}: We now show the effect of Dropconnect on the same network as mentioned in Theorem \ref{th:1}, where we multiply the individual elements of the weight matrix $\mathbf{W}_{\{k,k+1\}}$ with the elements of a matrix of i.i.d drawn Bernoulli selector random variables $\mathbf{U}^i_{\{k,k+1\}} \in \{0,1\}^{\{h_k,h_{k+1}\}}$ for all $k \in \{0,\ldots,P\}$ and all samples $i \in \{1,\ldots,m\}$. The elements of the matrix $\mathbf{U}^i_{\{k,k+1\}}$ are the vectors $\mathbf{u}^{\{i,t\}}_{\{k,k+1\}} \in \{0,1\}^{h_k},\; \forall\; t \in \{1,\ldots,h_{k+1}\}$ and each vector is composed of Bernoulli random variable such that $u^{\{i,n,t\}}_{\{k,k+1\}} = \delta^{\{i,n,t\}}_{\{k,k+1\}},\; \forall\; n \in \{1,\ldots,h_k\}$, which is $1$ with probability $p_{\{k,k+1\}}$ and $0$ with probability $q_{\{k,k+1\}} = 1 - p_{\{k,k+1\}}$. The hypothesis class of feed-forward neural networks with Dropconnect regularizer is given by:
\begin{gather}
\begin{split}
\mathcal{F}_{dc} = \sigma(\sigma(\ldots \sigma(\mathbf{x} \cdot (\mathbf{U}_{\{0,1\}} \odot \mathbf{W}_{\{0,1\}})) \cdot (\mathbf{U}_{\{1,2\}} \odot
\mathbf{W}_{\{1,2\}}) \ldots \cdot (\mathbf{U}_{\{P-2,P-1\}} \odot \mathbf{W}_{\{P-2,P-1\}})) \\\cdot (\mathbf{U}_{\{P-1,P\}} \odot \mathbf{W}_{\{P-1,P\}}))\cdot (\mathbf{u}^1_{\{P,P+1\}} \odot \mathbf{w}_{\{P,P+1\}}^1) 
\end{split} \nonumber
\label{eq:F-dropconnect}
\end{gather}
Here, $\odot$ represents element wise multiplication.

\begin{theorem}\label{th:3}
	For the same network as mentioned in Theorem \ref{th:1}, when added with Dropconnect to each layer $k$ for all  $k \in \{1,\ldots,P+1\}$ with Dropconnect probability $q_{\{k,k+1\}} = 1 - p_{\{k,k+1\}}$, the VC dimension bound $VCdim(\mathcal{F}_{dc})$ is bounded by:
	\begin{gather}
	VCdim(\mathcal{F}_{dc}) \leq p_{\{0,1\}} R^2A_{P+1}^2 L_\sigma^{2P}\prod_{k=1}^{P} p_{\{k,k+1\}}h_k A_k^2 \label{eq:VC FNN_DC1}
	\end{gather}
\end{theorem}
\textbf{Implications of Dropout and Dropconnect}: The two bounds presented in eq. \ref{eq:VC FNN_DO1} and eq. \ref{eq:VC FNN_DC1} are equivalent. Thus the two techniques brings down the capacity of the network, thus preventing problems like overfitting. The reasons as to why these two methods outperform other kinds of regularizers is they act like ensemble of networks and allow to learn representations from fewer number of neurons or weights at each iterations. Details of such an interpretation is mentioned in \cite{srivastava2014dropout} and \cite{wan2013regularization}.\\

\textbf{Bounds for a Resnet architecture}: Consider a generic resnet with $T$ residual blocks having $\acute{T}$ residual units per block. Each of the residual unit consists of activation function $\sigma(\cdot)$ followed by convolution layer (\textit{cv}), followed by dropout, $\sigma(\cdot)$ and \textit{cv} layer. The final output of \textit{cv} layer is added with the output of previous layer. We use a \textit{cv} layer ($cv_0(\cdot)$) after the input to increase the number of filters. After every one resnet block, we have a \textit{cv} layer, \textit{max-pool} or an \textit{average-pool} layer for dimensionality reduction. For our discussions, we keep a \textit{cv} unit $cv_i,\; \forall\; i \in \{0,\ldots,T\}$ for dimensionality reduction rather than \textit{max-pool} or \textit{average-pool}. After $T$ residual blocks we have $P$ fully connected layers with dropout. Lastly, we have our classifier layer and the hinge loss is applied to the classifier layer.
 
Consider the input data $\mathbf{x}^i = \phi_{0}(\mathbf{x}^i) \in \Re^{\{h_0^0,h_0^1,h_0^2\}}$. Let the number of filters in each \textit{cv} layer in block $r$ be $N_r,\; \forall\; r \;\in \{1,\ldots,R\}$ and size of filters for the \textit{cv} layers in those blocks as well as $cv_i$ dimensionality reduction blocks be $\{v_{r}^0 \times v_{r}^1\}$ with strides $\{s_{r}^0 \times s_{r}^1\}$. 
The function $cv(\cdot)$ takes in filter size, number of filters, strides and padding as the parameters alongside the input, which are not shown for brevity. The output of residual unit $\acute{r}$ in block $r$ is given by:
\begin{gather}
\phi_{\{r,\acute{r}\}}(\mathbf{x}^i) = \phi_{\{r,\acute{r}-1\}}(\mathbf{x}^i) + cv_{\{r,\acute{r},2\}}(\sigma(\mathbf{U}_{r}\odot cv_{\{r,\acute{r},1\}}(\sigma(\phi_{\{r,\acute{r}-1\}}(\mathbf{x}^i))))),\; \forall\; r,\acute{r} \in \{2,\ldots,T\}\times\{2,\ldots,\acute{T}\} \nonumber\\
\phi_{\{r,1\}}(\mathbf{x}^i) = cv_{r-1}(\phi_{\{r-1,\acute{R}\}}(\mathbf{x}^i)) + cv_{\{r,1,2\}}(\sigma(\mathbf{U}_{r}\odot cv_{\{r,1,1\}}(\sigma(cv_{r-1}(\phi_{\{r-1,\acute{R}\}}(\mathbf{x}^i)))))),\; \forall\; r \in \{2,\ldots,T\}\nonumber\\
\phi_{\{1,1\}}(x^i) = cv_0(\phi_{0}(\mathbf{x}^i)) + cv_{\{1,\acute{r},2\}}(\sigma(\mathbf{U}_{1}\odot cv_{\{1,\acute{r},1\}}(\sigma(cv_0(\phi_{0}(\mathbf{x}^i))))))\nonumber
\end{gather}


\begin{theorem}\label{th:4}
	The $VCdim(\mathcal{F})$ bound of a residual network as described above is given by:
	\begin{multline}
	VCdim(\mathcal{F})\leq R^2A_{P+1}^2\bigg[\left(L_\sigma^{2P}\prod_{k=1}^{P} p_kh_k A_{k}^2\right)\bigg(\left(A_{0}N_{0}v_0^2\right)\prod_{r=1}^{T}\left(A_rN_rv_r^2\right)^{3\acute{T}}L_{\sigma}^{4\acute{T}}p_r^{\acute{T}} \bigg)\bigg] \label{eq:VC RES1}
	\end{multline} 	
\end{theorem}
\textbf{Implications of the VC bound on Resnet architecture}: The bound given in eq. \ref{eq:VC RES1} is dependent on the max-norm of the weights, size of filters in each block, dropout probability, number of blocks, residual units per block and the Lipschitz constant of the activation function. It shows that the bound increases exponentially in the number of residual units per block which is expected as number of residual units increases the capacity of the network.

\subsection{Robustness to input noise}
Robustness measures the variation of the loss function w.r.t. the input $(x, y) \sim \mathcal{D}$. \citep{xu2012robustness} presented generalization bounds for an algorithm $\mathcal{A}$ being $\epsilon(\cdot)$ robust in terms of Radamacher averages. The idea that a large margin implying robustness was applied to deep networks in \citep{sokolic2017robust}. Here, we present the idea of robustness of an algorithm in terms of the VC dimension by incorporating the notion noise in a sample such that its label remains unchanged. Theorem \ref{th:5} shows that for a robust algorithm the VC dimension is larger than a non-robust algorithm.

\begin{theorem}\label{th:5}
	Consider the set $\mathcal{T} = \bigg\{(\Delta^1,\ldots,\Delta^m)  | \max_i \|\Delta^i\|_2 \leq c,\; \forall\; i \in \{1,\ldots,m\}\bigg\}$. Let this set denote the noise that can be added to the samples $\mathbf{x}^i$ to obtain $\mathbf{\hat{x}}^i = \mathbf{x}^i + \Delta^i$ such that for some $f \in \mathcal{F}_{ro}$, $f(\hat{\mathbf{x}}^i) = f(x^i),\; \forall\; i \in \{1,\ldots,m\}$, then the VC bound for such a hypothesis class $\mathcal{F}_{ro}$ is given by:
	\begin{gather}
	VCdim(\mathcal{F}_{ro}) \leq A_{P+1}^2 (R^2+c^2) L_\sigma^{2P}\prod_{k=1}^{P} h_k A_{k}^2  \nonumber
	\end{gather}
\end{theorem}
\textbf{Gradient regularization}: Consider the input noise-robust loss function,
\begin{gather}
\max_{(\Delta^1,\ldots,\Delta^m) \in \mathcal{T}} \sum_{i=1}^{m} \max \big(0, 1 - y_i(\phi_{P}(\mathbf{\hat{x}}^i)\cdot \mathbf{w}_{\{P,P+1\}}^1 )\big) \nonumber
\end{gather}
We now use the first order approximation for $\phi_{P}(\mathbf{\hat{x}}^i) = \phi_{P}(\mathbf{x^i} + \Delta^i)$ to get,
\begin{gather}
\phi_{P}(\mathbf{x^i} + \Delta^i) \approxeq \phi_{P}(\mathbf{x^i}) + \frac{\partial \phi_{P}(\mathbf{x^i})}{\partial \mathbf{x^i}} \cdot \Delta^i \label{eq: approx_phi1}
\end{gather}
Using eq. \ref{eq: approx_phi1} the objective function can be written as:
\begin{gather}
\max_{(\Delta^1,\ldots,\Delta^m) \in \mathcal{T}} \sum_{i=1}^{m} \max \big(0, 1 - y_i(\phi_{P}(\mathbf{x^i}) + \frac{\partial \phi_{P}(\mathbf{x^i})}{\partial \mathbf{x^i}} \cdot \Delta^i)\cdot \mathbf{w}_{\{P,P+1\}}^1 \big)\nonumber\\
\leq  \sum_{i=1}^{m} \max \big(0, 1 - y_i(\phi_{P}(\mathbf{x^i})\cdot \mathbf{w}_{\{P,P+1\}}^1 )\big) + \max_{(\Delta^1,\ldots,\Delta^m) \in \mathcal{T}} \sum_{i=1}^{m}\bigg(\frac{\partial \phi_{P}(\mathbf{x^i})}{\partial \mathbf{x^i}} \cdot \Delta^i\bigg)\cdot \mathbf{w}_{\{P,P+1\}}^1 \nonumber\\
\leq  \sum_{i=1}^{m} \max \big(0, 1 - y_i(\phi_{P}(\mathbf{x^i})\cdot \mathbf{w}_{\{P,P+1\}}^1 )\big) + \max_{(\Delta^1,\ldots,\Delta^m) \in \mathcal{T}} \sum_{i=1}^{m}\left\Vert\bigg(\frac{\partial \phi_{P}(\mathbf{x^i})}{\partial \mathbf{x^i}} \cdot \Delta^i\bigg)\right\Vert_2\Vert\mathbf{w}_{\{P,P+1\}}^1\Vert_2 \nonumber\\
\leq  \sum_{i=1}^{m} \max \big(0, 1 - y_i(\phi_{P}(\mathbf{x^i})\cdot \mathbf{w}_{\{P,P+1\}}^1 )\big) + \max_{(\Delta^1,\ldots,\Delta^m) \in \mathcal{T}} \sum_{i=1}^{m}\left\Vert \frac{\partial \phi_{P}(\mathbf{x^i})}{\partial \mathbf{x^i}} \right\Vert_2 \left\Vert \Delta^i \right\Vert_2 A_{P+1} \nonumber\\
\leq  \sum_{i=1}^{m} \max \big(0, 1 - y_i(\phi_{P}(\mathbf{x^i})\cdot \mathbf{w}_{\{P,P+1\}}^1 )\big) +  \sum_{i=1}^{m}\left\Vert \frac{\partial \phi_{P}(\mathbf{x^i})}{\partial \mathbf{x^i}} \right\Vert_F c A_{P+1} \label{eq: grad_a1}
\end{gather}

The term $\left\Vert \frac{\partial \phi_{P}(\mathbf{x^i})}{\partial \mathbf{x^i}} \right\Vert_2$ is the norm of Jacobian matrix $J(\mathbf{x})$ of the deep neural network (DNN). We now show that minimizing the term is equivalent to maximizing the input margin of the DNN. 
\textbf{Input and output margin}: The input margin of sample $\mathbf{x^i}$ can be defined as:
\begin{gather}
	\gamma_{ip}^i = \sup_a\{\left\Vert \mathbf{x}^i-\mathbf{x}\right\Vert_2 \leq a;\; sign(\phi_{P+1}(\mathbf{x})) =  sign(\phi_{P+1}(\mathbf{x}^i))\} \label{eq: ip_mar1}
\end{gather}
whereas, the output margin of the sample $\mathbf{x^i}$ is given as:
\begin{gather}
\gamma_{op}^i = \sup_a\{\left\Vert \phi_{P}(\mathbf{x}^i)-\phi_{P}(\mathbf{x})\right\Vert_2 \leq a;\; sign(\phi_{P+1}(\mathbf{x})) =  sign(\phi_{P+1}(\mathbf{x}^i))\} \label{eq: op_mar1}
\end{gather}
Using the Theorem 3, Corollary 2 of \cite{sokolic2017robust} and the Lebesgue differentiation theorem, we get,
\begin{gather}
\left\Vert \phi_{P}(\mathbf{x}^i)-\phi_{P}(\mathbf{x})\right\Vert_2 \leq \sup_{\mathbf{x^i},\mathbf{x},\;t \in [0,1]}\left\Vert J\big(\mathbf{x} + t(\mathbf{x^i}-\mathbf{x})\big)\right\Vert_2 \left\Vert \mathbf{x^i} - \mathbf{x}\right\Vert_2 \label{eq: grad1}
\end{gather}
Assume that the point $\mathbf{x}$ lies on the decision boundary, then the term $\left\Vert \mathbf{x^i} - \mathbf{x}\right\Vert_2 $ is equal to $\gamma_{ip}^i$. Using the aforementioned fact, one can write:
\begin{gather}
\sup_{\mathbf{x^i},\mathbf{x},t \in [0,1]}\left\Vert J\big(\mathbf{x} + t(\mathbf{x^i}-\mathbf{x})\big)\right\Vert_2 = \sup_{\mathbf{x}: \left\Vert \mathbf{x^i} - \mathbf{x}\right\Vert_2 \leq \gamma_{ip}^i}\left\Vert J(\mathbf{x})\right\Vert_2 \label{eq: grad2}
\end{gather}
Using eqs. \ref{eq: grad2}, \ref{eq: ip_mar1} and \ref{eq: op_mar1} in eq. \ref{eq: grad1} we get,
\begin{gather}
\gamma_{op}^i \leq  \sup_{\mathbf{x}: \left\Vert \mathbf{x^i} - \mathbf{x}\right\Vert_2 \leq \gamma_{ip}^i}\left\Vert J(\mathbf{x})\right\Vert_2 \gamma_{ip}^i \label{eq: grad3}\nonumber\\
\implies \gamma_{ip}^i \geq \frac{\gamma_{op}^i}{\sup_{\mathbf{x}: \left\Vert \mathbf{x^i} - \mathbf{x}\right\Vert_2 \leq \gamma_{ip}^i}\left\Vert J(\mathbf{x})\right\Vert_2} \label{eq: grad4}
\end{gather}
Let $conv(\mathcal{X}) = \{\mathbf{x} : \mathbf{x} + t(\mathbf{x^i}-\mathbf{x}) ,\; t \in [0,1] \} $ denotes convex set, then from eq. \ref{eq: grad1} and eq. \ref{eq: grad4} we can write,
\begin{gather}
\gamma_{ip}^i \geq \frac{\gamma_{op}^i}{\sup_{\mathbf{x} \in conv(\mathcal{X})} \left\Vert J(\mathbf{x})\right\Vert_2}\label{eq: grad5}
\end{gather}
From eq. \ref{eq: grad5} we see that, minimizing the norm of Jacobian matrix $J(\mathbf{x})$ amounts to increasing the lower bound on input margin whereas, eq. \ref{eq: grad_a1} shows that minimizing the norm of $J(\mathbf{x})$ along with the hinge loss approximates the input noise-robust hinge loss function. The two hints at the fact that maximizing the input margin is required to obtain an input noise-robust deep architecture.
\section{Conclusion}
\label{sec:conclusion}
This paper studies the radius margin bounds for deep architectures both fully connected and residual convolutional networks in presence of hinge loss at the output. We show that the capacity of the deep architecture can be bounded by the number of neurons, the filter size for each layer, the Dropout probability or Dropconnect probability and the max norm of the weights. 
We also hint at the equivalence of minimizing the norm of the Jacobian matrix of the network and robustness to the input perturbation. We show that minimizing the norm of the Jacobian leads to a network with large input margin which in turn causes the network to be robust to perturbation in the input space. In the future, we would like to study the effect of weight quantization on the VC dimension bound of the deep architectures.

\acks{We would like to acknowledge support for this project	from Safran Group.}

\newpage
\appendix
\section*{Appendix} \label{Appendix}
\label{sec:appendix}
\subsection*{Proof of Theorem \ref{th:1}} \label{pr:1}
\textbf{Proof}: 
Since the set $\{\mathbf{x}^1,\ldots,\mathbf{x}^m\}$ is fully shattered by the hypothesis class, implies for all $\mathbf{y} = (y_1,\ldots,y_m) \in \{-1,1\}^m$, there exists $\mathbf{w}_{\{P,P+1\}}^1$, such that,
\begin{gather}
\forall \; i \in [1,m],\;\; 1 \leq y_i(\phi_P(\mathbf{x}^i) \cdot \mathbf{w}_{\{P,P+1\}}^1) \label{eq: error1}
\end{gather}
Summing up these inequalities yields,
\begin{gather}
m \leq  \big(\sum_{i=1}^{m} y_i \phi_P(\mathbf{x}^i)\big) \cdot \mathbf{w}_{\{P,P+1\}}^1  \leq \|w_{\{P,P+1\}}^1 \|_2 \|\sum_{i=1}^{m} y_i \phi_P(\mathbf{x}^i)\|_2 \leq A_{P+1} \|\sum_{i=1}^{m} y_i \phi_P(\mathbf{x}^i)\|_2 \nonumber
\end{gather}
Since, the inequality holds for all $\mathbf{y} \in \{-1,1\}^m$, it also holds on expectation over $(y_1,\ldots,y_m)$ drawn i.i.d. according to a uniform distribution over $\{-1,1\}$. Since the distribution is uniform, hence for $i \neq j$, we have $E[y_iy_j] = E[y_i]E[y_j]$. Thus, since the distribution is uniform $E[y_iy_j] = 0$ if $i \neq j$, $E[y_iy_j] = 1$ otherwise. This gives,
\begin{gather}
m \leq A_{P+1} E_{\mathbf{y}}[\|\sum_{i=1}^{m} y_i \phi_P(\mathbf{x}^i)\|_2] \nonumber
\end{gather}
Applying Jensen's inequality,
\begin{gather}
  m \leq A_{P+1} \big[E_{\mathbf{y}}[\|\sum_{i=1}^{m} y_i \phi_P(\mathbf{x}^i)\|_2^2] \big]^{\frac{1}{2}} \nonumber\\
  = A_{P+1} \big[\sum_{i=1}^{m} \sum_{j=1}^{m} E_{\mathbf{y}}[y_i y_j] (\phi_P(\mathbf{x}^i) \cdot \phi_P(\mathbf{x}^j)) \big]^{\frac{1}{2}}\nonumber\\
  = A_{P+1} \big[\sum_{i=1}^{m} (\phi_P(\mathbf{x}^i) \cdot \phi_P(\mathbf{x}^i)) \big]^{\frac{1}{2}} \nonumber\\
  \leq A_{P+1} \big[m \max_i \|\phi_P(\mathbf{x}^i)\|_2^2 \big]^{\frac{1}{2}} \nonumber\\
  \implies \sqrt{m} \leq A_{P+1} \big[ \max_i \|\phi_P(\mathbf{x}^i)\|_2^2 \big]^{\frac{1}{2}} \label{eq: VC FNN3}
\end{gather}
Now, we prove the bound on $r = \max_i \|\phi_P(\mathbf{x}^i)\|_2^2$.
\begin{gather}
r  = \max_i \|[\sigma(\phi_{P-1}(\mathbf{x^i}) \cdot \mathbf{w}_{\{P-1,P\}}^1),\ldots,\sigma(\phi_{P-1}(\mathbf{x^i})\cdot \mathbf{w}_{\{P-1,P\}}^{h_P})]\|_2^2\nonumber
\end{gather}
Let $t$ be the index of the maximum absolute value in the vector $\phi_P(\mathbf{x}^i)$
\begin{gather}
  \leq \max_i h_P\|\sigma(\phi_{P-1}(\mathbf{x^i}) \cdot \mathbf{w}_{\{P-1,P\}}^t)\|^2\nonumber
\end{gather}  
Using eq. \ref{eq:lipschitz_norm} we get,
\begin{gather}
  \leq \max_i h_P L_\sigma^2 \|\phi_{P-1}(\mathbf{x^i}) \|_2^2\|\mathbf{w}_{\{P-1,P\}}^t\|_2^2\nonumber\\
  = h_P A_{P}^2 L_\sigma^2 \max_i \|\phi_{P-1}(\mathbf{x^i})\|_2^2 \nonumber
\end{gather}
Applying it recursively till layer 1 we get,  
\begin{gather}
  \leq (\max_i \|\phi_{0}(\mathbf{x^i})\|_2^2 ) L_\sigma^{2P}\prod_{k=1}^{P} h_k A_{k}^2 \nonumber\\
  = (\max_i \|\mathbf{x^i}\|_2^2 ) L_\sigma^{2P}\prod_{k=1}^{P} h_k A_{k}^2 \label{eq: VC FNN3b} \\
\implies r \leq R^2 L_\sigma^{2P}\prod_{k=1}^{P} h_k A_{k}^2 \label{eq: VC FNN4}
\end{gather}
using eq. \ref{eq: VC FNN4} in eq. \ref{eq: VC FNN3} we get,
\begin{gather}
\sqrt{m} \leq A_{P+1} \big[R^2 L_\sigma^{2P}\prod_{k=1}^{P} h_k A_{k}^2 \big]^{\frac{1}{2}} \nonumber\\
m \leq A_{P+1}^2 R^2 L_\sigma^{2P}\prod_{k=1}^{P} h_k A_{k}^2  \nonumber\\
\implies VCdim(\mathcal{F}) \leq A_{P+1}^2 R^2 L_\sigma^{2P}\prod_{k=1}^{P} h_k A_{k}^2  \nonumber
\end{gather}

\subsection*{Proof of Theorem \ref{th:2}} \label{pr:2}
\textbf{Proof}: Since, $\mathbf{u}_k^i$ is a vector of random variables for each sample $i$ and each layer $k$, we have to take expectations over each random variable present to determine the expected VC dimension of the network. Here, $\phi_k(\mathbf{x}^i) = \sigma((\mathbf{u}_{k-1}^i \odot \phi_{k-1}(\mathbf{x}^i))\cdot \mathbf{W}_{\{k-1,k\}})$. Following eq. \ref{eq: error1} we get,
\begin{gather}
\forall \; i \in [1,m],\;\; 1 \leq y_i(\mathbf{u}_{P}^i \odot \phi_P(\mathbf{x}^i) \cdot \mathbf{w}_{\{P,P+1\}}^1) \label{eq: error2}
\end{gather}
Since, the inequality holds for all $y_i \in \{-1,1\}$ and for all $\mathbf{u}_{P}^i$, it also holds on expectation for $\{y_1,\ldots,y_m\}$  and $\mathbf{U}_{k} = \{\mathbf{u}_{k}^1,\ldots,\mathbf{u}_{k}^m\},\; \forall\; k \in \{0,\ldots,P\}$. The distribution over $\mathbf{u}_{k}^i$ is Bernoulli, thus $E[\mathbf{u}_{\{k,t\}}^i] = p_k$. This gives,
\begin{gather}
m \leq A_{P+1} E_{\mathbf{y}}E_{\mathbf{U}_{0},\ldots,\mathbf{U}_{P}}[\|\sum_{i=1}^{m} y_i \mathbf{u}_{P}^i \odot \phi_P(\mathbf{x}^i)\|_2]\nonumber
\end{gather}
Applying Jensen's inequality,
\begin{gather}
m \leq A_{P+1} \big[E_{\mathbf{y}}E_{\mathbf{U}_{0},\ldots,\mathbf{U}_{P}}[\|\sum_{i=1}^{m} y_i \mathbf{u}_{P}^i \odot \phi_P(\mathbf{x}^i)\|_2^2] \big]^{\frac{1}{2}}\nonumber\\
= A_{P+1} \big[\sum_{i=1}^{m} \sum_{j=1}^{m} E_{\mathbf{y}}[y_i y_j] E_{\mathbf{U}_{0},\ldots,\mathbf{U}_{P}} [ (\mathbf{u}_{P}^i \odot \phi_P(\mathbf{x}^i)) \cdot (\mathbf{u}_{P}^j \odot \phi_P(\mathbf{x}^j))] \big]^{\frac{1}{2}}\nonumber\\
= A_{P+1} \big[\sum_{i=1}^{m} E_{\mathbf{U}_{0},\ldots,\mathbf{U}_{P}}[ (\mathbf{u}_{P}^i \odot \phi_P(\mathbf{x}^i)) \cdot (\mathbf{u}_{P}^i \odot  \phi_P(\mathbf{x}^i))] \big]^{\frac{1}{2}} \nonumber\\
= A_{P+1} \bigg[\sum_{i=1}^{m} E_{\mathbf{U}_{0},\ldots,\mathbf{U}_{P-1}}E_{\mathbf{U}_{P}}\big[\; \big((\delta_{\{P,1\}}^i)^2 \phi_{P}(\mathbf{x}_{1}^i)^2 + \ldots + (\delta_{\{P,h_P\}}^i)^2 \phi_{P}(\mathbf{x}_{h_P}^i)^2\big)\; \big] \bigg]^{\frac{1}{2}} \nonumber
\end{gather}
Using the fact that $E[(\delta_{\{P,t\}}^i)^2] = E[(\delta_{\{P,t\}}^i)] = p_P$ we get,
\begin{gather}
= A_{P+1} \bigg[\sum_{i=1}^{m} p_P  E_{\mathbf{U}_{0},\ldots,\mathbf{U}_{P-1}} [\|\phi_P(\mathbf{x}^i)\|_2^2]\bigg]^{\frac{1}{2}} \nonumber\\
\leq A_{P+1} \bigg[p_P m  \max_i E_{\mathbf{U}_{0},\ldots,\mathbf{U}_{P-1}}[\|\phi_P(\mathbf{x}^i)\|_2^2]\bigg]^{\frac{1}{2}} \nonumber\\
\implies \sqrt{m} \leq A_{P+1} \bigg[p_P \max_i E_{\mathbf{U}_{0},\ldots,\mathbf{U}_{P-1}}[\|\phi_P(\mathbf{x}^i)\|_2^2]\bigg]^{\frac{1}{2}} \label{eq: VC FNN5}
\end{gather}

Now, we prove the bound on $r = \max_i E_{\mathbf{U}_{0},\ldots,\mathbf{U}_{P-1}} [\|\phi_P(\mathbf{x}^i)\|_2^2]$.
\begin{gather}
r= \max_i E_{\mathbf{U}_{0},\ldots,\mathbf{U}_{P-1}}  [\|[\sigma( (\mathbf{u}_{P-1}^i \odot \phi_{P-1}(\mathbf{x^i})) \cdot \mathbf{w}_{\{P-1,P\}}^1),\ldots,(\mathbf{u}_{P-1}^i \odot \sigma(\phi_{P-1}(\mathbf{x^i}))\cdot \mathbf{w}_{\{P-1,P\}}^{h_P})]\|_2^2]\nonumber
\end{gather}
Let $t$ be the index of the maximum absolute value in the vector $\phi_P(\mathbf{x}^i)$ 
\begin{gather}
\leq h_P \max_i E_{\mathbf{U}_{0},\ldots,\mathbf{U}_{P-1}}  [\|\sigma((\mathbf{u}_{P-1}^i \odot \phi_{P-1}(\mathbf{x^i})) ) \cdot \mathbf{w}_{\{P-1,P\}}^t\|^2]\nonumber
\end{gather}
Using eq. \ref{eq:lipschitz_norm} we get,
\begin{gather}
\leq  h_P L_\sigma^2 \max_i E_{\mathbf{U}_{0},\ldots,\mathbf{U}_{P-2}} E_{\mathbf{U}_{P-1}}[\|\mathbf{u}_{P-1}^i \odot \phi_{P-1}(\mathbf{x^i})\|_2^2]\|\mathbf{w}_{\{P-1,P\}}^t\|_2^2\nonumber\\
= h_P A_{P}^2 L_\sigma^2 \max_i  E_{\mathbf{U}_{0},\ldots,\mathbf{U}_{P-2}} E_{\mathbf{U}_{P-1}} [\|\mathbf{u}_{P-1}^i \odot \phi_{P-1}(\mathbf{x^i})\|_2^2] \nonumber
\end{gather}
Applying the expectation over $\mathbf{u}_{P-1}^i$ we get,
\begin{gather}
= h_P A_{P}^2 L_\sigma^2 p_{P-1}\max_i  E_{\mathbf{U}_{0},\ldots,\mathbf{U}_{P-2}}[\| \phi_{P-1}(\mathbf{x^i})\|_2^2] \nonumber
\end{gather}
Applying it recursively till layer 1 we get,
\begin{gather}
\leq L_\sigma^{2P}\prod_{k=1}^{P} p_kh_k A_{k}^2 (\max_i  E_{\mathbf{U}_{0}}[\|\phi_{0}(\mathbf{x^i})\|_2^2 )] \label{eq: dropout1} \\
= L_\sigma^{2P} p_0\prod_{k=1}^{P} p_kh_k A_{k}^2 (\max_i \|\mathbf{x^i}\|_2^2 )  \nonumber\\
\implies r \leq R^2 p_0L_\sigma^{2P}\prod_{k=1}^{P} p_kh_k A_{k}^2 \label{eq: VC FNN6}
\end{gather}
using eq. \ref{eq: VC FNN6} in eq. \ref{eq: VC FNN5} we get,
\begin{gather}
\sqrt{m} \leq A_{P+1} \big[R^2p_0L_\sigma^{2P}\prod_{k=1}^{P}p_kh_k A_{k}^2 \big]^{\frac{1}{2}} \nonumber\\
m \leq A_{P+1}^2 R^2 p_0 L_\sigma^{2P}\prod_{k=1}^{P} p_k h_k A_{k}^2  \nonumber\\
\implies VCdim(\mathcal{F}_{do}) \leq A_{P+1}^2 R^2 p_0 L_\sigma^{2P}\prod_{k=1}^{P} p_k h_k A_{k}^2  \nonumber
\end{gather}

\subsection*{Proof of Theorem \ref{th:3}} \label{pr:3}
\textbf{Proof}: Following eq. \ref{eq: error1} we get,
\begin{gather}
\forall \; i \in [1,m],\;\; 1 \leq y_i(\phi_P(\mathbf{x}^i) \cdot (\mathbf{u}_{\{P,P+1\}}^{\{i,1\}} \odot\mathbf{w}_{\{P,P+1\}}^1)) \nonumber\\
\implies m \leq \sum_{i=1}^{m} y_i(\phi_P(\mathbf{x}^i) \cdot (\mathbf{u}_{\{P,P+1\}}^{\{i,1\}} \odot\mathbf{w}_{\{P,P+1\}}^1))\nonumber\\
           = \sum_{i=1}^{m} y_i( (\mathbf{u}_{\{P,P+1\}}^{\{i,1\}}\odot \phi_P(\mathbf{x}^i)) \cdot \mathbf{w}_{\{P,P+1\}}^1) \label{eq: error3}
\end{gather}
Since, the inequality holds for all $y_i \in \{-1,1\}$ and for all $\mathbf{u}_{\{P,P+1\}}^i$, it also holds on expectation for $\{y_1,\ldots,y_m\}$  and $\mathbf{U'}_{\{k,k+1\}} = \{\mathbf{U}_{\{k,k+1\}}^1,\ldots,\mathbf{U}_{\{k,k+1\}}^m\},\; \forall\; k \in \{0,\ldots,P\}$. The elements of the vector $\mathbf{u}_{\{k,k+1\}}^{\{i,t\}}, \;\forall \; t \in \{1,\ldots,h_{k+1}\}$ are distributed according to the Bernoulli distribution, thus $E[\mathbf{u}_{\{k,k+1\}}^{\{i,n,t\}}] = E[\delta_{\{k,k+1\}}^{\{i,n,t\}}] = p_{k,k+1},\; \forall\; n \in \{1,\ldots,h_k\}$. This gives,
\begin{gather}
m \leq A_{P+1} E_{\mathbf{y}}E_{\mathbf{U}_{\{0,1\}},\ldots,\mathbf{U}_{\{P,P+1\}}}[\|\sum_{i=1}^{m} y_i (\mathbf{u}_{\{P,P+1\}}^{\{i,1\}}\odot \phi_P(\mathbf{x}^i))\|_2]\nonumber
\end{gather}
Using Jensen's inequality,
\begin{gather}
\leq A_{P+1} \big[E_{\mathbf{y}}E_{\mathbf{U}_{\{0,1\}},\ldots,\mathbf{U}_{\{P,P+1\}}}[\|\sum_{i=1}^{m} y_i (\mathbf{u}_{\{P,P+1\}}^{\{i,1\}}\odot \phi_P(\mathbf{x}^i))\|_2^2] \big]^{\frac{1}{2}}\nonumber\\
= A_{P+1} \big[\sum_{i=1}^{m} \sum_{j=1}^{m} E_{\mathbf{y}}[y_i y_j] E_{\mathbf{U}_{\{0,1\}},\ldots,\mathbf{U}_{\{P,P+1\}}} [(\mathbf{u}_{\{P,P+1\}}^{\{i,1\}} \odot \phi_P(\mathbf{x}^i)) \cdot (\mathbf{u}_{\{P,P+1\}}^{\{j,1\}} \odot \phi_P(\mathbf{x}^j))] \big]^{\frac{1}{2}}\nonumber\\
= A_{P+1} \big[\sum_{i=1}^{m} E_{\mathbf{U}_{\{0,1\}},\ldots,\mathbf{U}_{\{P,P+1\}}}[ (\mathbf{u}_{\{P,P+1\}}^{\{i,1\}} \odot \phi_P(\mathbf{x}^i)) \cdot (\mathbf{u}_{\{P,P+1\}}^{\{i,1\}} \odot \phi_P(\mathbf{x}^i))] \big]^{\frac{1}{2}} \nonumber\\
= A_{P+1} \bigg[\sum_{i=1}^{m}E_{\mathbf{U}_{\{0,1\}},\ldots,\mathbf{U}_{\{P-1,P\}}}E_{\mathbf{U}_{\{P,P+1\}}}\big[\; \big((\delta_{\{P,P+1\}}^{\{i,1,1\}})^2 \phi_{P}(\mathbf{x}_{1}^i)^2 + \ldots + (\delta_{\{P,P+1\}}^{{\{i,h_P,1\}}})^2 \phi_{P}(\mathbf{x}_{h_P}^i)^2\big)\; \big] \bigg]^{\frac{1}{2}}\nonumber
\end{gather}
Using the fact that $E[(\delta_{\{P,P+1\}}^{\{i,n,1\}})^2] = E[(\delta_{\{P,P+1\}}^{\{i,n,1\}})] = p_{\{P,P+1\}}$ we get,
\begin{gather}
= A_{P+1} \bigg[\sum_{i=1}^{m} p_{P,P+1} E_{\mathbf{U}_{\{0,1\}},\ldots,\mathbf{U}_{\{P-1,P\}}} [\|\phi_P(\mathbf{x}^i)\|_2^2]\bigg]^{\frac{1}{2}}\nonumber \\
\leq A_{P+1} \bigg[p_{\{P,P+1\}} m  \max_i E_{\mathbf{U}_{\{0,1\}},\ldots,\mathbf{U}_{\{P-1,P\}}}[\|\phi_P(\mathbf{x}^i)\|_2^2]\bigg]^{\frac{1}{2}}\nonumber \\
\implies \sqrt{m} \leq A_{P+1} \bigg[p_{\{P,P+1\}} \max_i E_{\mathbf{U}_{\{0,1\}},\ldots,\mathbf{U}_{\{P-1,P\}}}[\|\phi_P(\mathbf{x}^i)\|_2^2]\bigg]^{\frac{1}{2}} \label{eq: VC FNN7}
\end{gather}

Now, we prove the bound on $r = \max_i E_{\mathbf{U}_{\{0,1\}},\ldots,\mathbf{U}_{\{P-1,P\}}}[\|\phi_P(\mathbf{x}^i)\|_2^2]$.
\begin{gather}
\begin{split}
r= \max_i E_{\mathbf{U}_{\{0,1\}},\ldots,\mathbf{U}_{\{P-1,P\}}}  [\|[\sigma(\phi_{P-1}(\mathbf{x^i}) \cdot (\mathbf{u}_{\{P,P+1\}}^{\{i,1\}} \odot \mathbf{w}_{\{P-1,P\}}^1)),\ldots,\\ \sigma(\phi_{P-1}(\mathbf{x^i}) \cdot (\mathbf{u}_{\{P,P+1\}}^{\{i,h_P\}} \odot \mathbf{w}_{\{P-1,P\}}^{h_P}))]\|_2^2]
\end{split}\nonumber
\end{gather}
Let $t$ be the index of the maximum absolute value in the vector $\phi_P(\mathbf{x}^i)$ 
\begin{gather}
\leq h_P \max_i E_{\mathbf{U}_{\{0,1\}},\ldots,\mathbf{U}_{\{P-1,P\}}} [\|\sigma(\phi_{P-1}(\mathbf{x^i}) \cdot (\mathbf{u}_{\{P,P+1\}}^{\{i,t\}} \odot \mathbf{w}_{\{P-1,P\}}^t))\|_2^2]\nonumber\\
\end{gather}
We use the Lipschitz property from eq. \ref{eq:lipschitz_norm}  and also the fact that $\phi_{P-1}(\mathbf{x^i}) \cdot (\mathbf{u}_{\{P,P+1\}}^{\{i,t\}}\odot \mathbf{w}_{\{P-1,P\}}^t) = (\mathbf{u}_{\{P,P+1\}}^{\{i,t\}} \odot \phi_{P-1}(\mathbf{x^i}))\cdot \mathbf{w}_{\{P-1,P\}}^t)$ to obtain,
\begin{gather}
\leq  h_P L_\sigma^2 \max_i E_{\mathbf{U}_{\{0,1\}},\ldots,\mathbf{U}_{\{P-2,P-1\}}} E_{\mathbf{U}_{\{P-1,P\}}}[\|\mathbf{u}_{\{P,P+1\}}^{i,t} \odot \phi_{P-1}(\mathbf{x^i})\|_2^2]\|\mathbf{w}_{\{P-1,P\}}^t\|_2^2\nonumber\\
= h_P A_{P}^2 L_\sigma^2 \max_i  E_{\mathbf{U}_{\{0,1\}},\ldots,\mathbf{U}_{\{P-2,P-1\}}} E_{\mathbf{U}_{\{P-1,P\}}} [\|\mathbf{u}_{\{P,P+1\}}^{i,t} \odot \phi_{P-1}(\mathbf{x^i})\|_2^2] \nonumber
\end{gather}
Applying the expectation over $\mathbf{u}_{\{P,P+1\}}^{i,t}$ we get,
\begin{gather}
= h_P A_{P}^2 L_\sigma^2 p_{\{P-1,P\}}\max_i   E_{\mathbf{U}_{\{0,1\}},\ldots,\mathbf{U}_{\{P-2,P-1\}}}[\| \phi_{P-1}(\mathbf{x^i})\|_2^2] \nonumber
\end{gather}
Applying it recursively till layer 1 we get,
\begin{gather}
\leq L_\sigma^{2P}\prod_{k=1}^{P} p_{\{k,k+1\}}h_k A_{k}^2 (\max_i  E_{\mathbf{U}_{\{0,1\}}}[\|\phi_{0}(\mathbf{x^i})\|_2^2 )] \nonumber \\
= L_\sigma^{2P} p_{\{0,1\}}\prod_{k=1}^{P} p_{\{k,k+1\}} h_k A_{k}^2 (\max_i \|\mathbf{x^i}\|_2^2 )  \nonumber\\
\implies r \leq R^2 p_{\{0,1\}} L_\sigma^{2P}\prod_{k=1}^{P} p_{\{k,k+1\}}h_k A_{k}^2 \label{eq: VC FNN8}
\end{gather}
using eq. \ref{eq: VC FNN8} in eq. \ref{eq: VC FNN7} we get,
\begin{gather}
\sqrt{m} \leq A_{P+1} \big[R^2p_{\{0,1\}}L_\sigma^{2P}\prod_{k=1}^{P}p_{\{k,k+1\}} h_k A_{k}^2 \big]^{\frac{1}{2}}\nonumber \\
m \leq A_{P+1}^2 R^2 p_{\{0,1\}} L_\sigma^{2P}\prod_{k=1}^{P} p_{\{k,k+1\}} h_k A_{k}^2  \nonumber\\
\implies VCdim(\mathcal{F}_{dc}) \leq A_{P+1}^2 R^2 p_{\{0,1\}} L_\sigma^{2P}\prod_{k=1}^{P}  p_{\{k,k+1\}} h_k A_{k}^2  \nonumber
\end{gather}

\subsection*{Proof of Theorem \ref{th:4}} \label{pr:4}
\textbf{Proof}: Following eq. \ref{eq: error2} - eq. \ref{eq: dropout1}, for $P$ fully connected layers and a classifier layer we get,
\begin{gather}
\sqrt{m} \leq A_{P+1}\left[L_\sigma^{2P}\prod_{k=1}^{P} p_kh_k A_{k}^2 (\max_i  E_{\mathbf{U}_{1}},\ldots,E_{\mathbf{U}_{T}}[\|cv_T(\phi_{\{T,\acute{T}\}}(\mathbf{x^i}))\|_2^2 ] )\right]^\frac{1}{2} \label{eq: res1}\\
\end{gather}
Let ${\{t_1,t_2,t_3\}}$ denote the index of maximum value of $cv_T(\phi_{\{T,\acute{T}\}}(\mathbf{x^i}))$. Let $w$ be the weight connecting the previous layer to the current layer whose norm is bounded by $A_{T}$. The part of previous layer connected to the current layer via the weight $w$ is given by $\acute{\phi}_{\{T,\acute{T}\}}(\mathbf{x^i})$. Let the filter sizes remain the same and equal to $v_T$ for the block $T$. Thus we get,
\begin{equation}
\leq A_{P+1}\left[\left(L_\sigma^{2P}\prod_{k=1}^{P} p_kh_k A_{k}^2\right)\left(\left(A_{T}N_{T}v_T^2\right)\max_i  E_{\mathbf{U}_{1}},\ldots,E_{\mathbf{U}_{T}}[\|\acute{\phi}_{\{T,\acute{T}\}}(\mathbf{x^i})\|_2^2 ]\right)\right]^\frac{1}{2} \nonumber
\end{equation}
\begin{multline}
= A_{P+1}\bigg[\left(L_\sigma^{2P}\prod_{k=1}^{P} p_kh_k A_{k}^2\right)\bigg(\left(A_{T}N_{T}v_T^2\right)
\max_i  E_{\mathbf{U}_{1}},\ldots,E_{\mathbf{U}_{T}}\\
[\|\acute{\phi}_{\{T,\acute{T}-1\}}(\mathbf{x}^i) + \acute{cv}_{\{T,\acute{T},2\}}(\sigma(\mathbf{U}_{T}\odot cv_{\{T,\acute{T},1\}}(\sigma(\phi_{\{T,\acute{T}-1\}}(\mathbf{x}^i)))))\|_2^2 ]\bigg)\bigg]^\frac{1}{2} \nonumber
\end{multline} 
Again, let ${\{t_1,t_2,t_3\}}$ denote the index of maximum value of $\acute{\phi}_{\{T,\acute{T}\}}(\mathbf{x^i})$, then for the last $cv(\cdot)$ layer of resnet unit, we get,
\begin{multline}
= A_{P+1}\bigg[\left(L_\sigma^{2P}\prod_{k=1}^{P} p_kh_k A_{k}^2\right)\bigg(\left(A_{T}N_{T}v_T^2\right)\left(N_Tv_T^2\right)
\max_i  E_{\mathbf{U}_{1}},\ldots,E_{\mathbf{U}_{T}}\\
[\|\acute{\phi}_{\{T,\acute{T}-1\}}(\mathbf{x}^i)^{\{t_1,t_2,t_3\}} + (\acute{\sigma}(\mathbf{U}_{T}\odot cv_{\{T,\acute{T},1\}}(\sigma(\phi_{\{T,\acute{T}-1\}}(\mathbf{x}^i)))))\dot w\|_2^2 ]\bigg)\bigg]^\frac{1}{2} \nonumber
\end{multline} 
Neglecting the effect of $\acute{\phi}_{\{T,\acute{T}-1\}}(\mathbf{x}^i)^{\{t_1,t_2,t_3\}}$ on the sum, and considering the norm of the weights bounded by $A_T$ we get,
\begin{multline}
= A_{P+1}\bigg[\left(L_\sigma^{2P}\prod_{k=1}^{P} p_kh_k A_{k}^2\right)\bigg(\left(A_{T}N_{T}v_T^2\right)\left(A_TN_Tv_T^2\right)
\max_i  E_{\mathbf{U}_{1}},\ldots,E_{\mathbf{U}_{T}}\\
[\|\acute{\sigma}(\mathbf{U}_{T}\odot cv_{\{T,\acute{T},1\}}(\sigma(\phi_{\{T,\acute{T}-1\}}(\mathbf{x}^i))))\|_2^2 ]\bigg)\bigg]^\frac{1}{2} \label{eq: VCRES1}
\end{multline} 
Using the Lipschitzness of $\sigma(\cdot)$ we get,
\begin{multline}
= A_{P+1}\bigg[\left(L_\sigma^{2P}\prod_{k=1}^{P} p_kh_k A_{k}^2\right)\bigg(\left(A_{T}N_{T}v_T^2\right)\left(A_TN_Tv_T^2\right)L_{\sigma}^2
\max_i  E_{\mathbf{U}_{1}},\ldots,E_{\mathbf{U}_{T}}\\
[\|\acute{\mathbf{U}}_{T}\odot \acute{cv}_{\{T,\acute{T},1\}}(\sigma(\phi_{\{T,\acute{T}-1\}}(\mathbf{x}^i)))\|_2^2 ]\bigg)\bigg]^\frac{1}{2} \nonumber
\end{multline} 
We now take an expectation over the dropout variable,
\begin{multline}
= A_{P+1}\bigg[\left(L_\sigma^{2P}\prod_{k=1}^{P} p_kh_k A_{k}^2\right)\bigg(\left(A_{T}N_{T}v_T^2\right)\left(A_TN_Tv_T^2\right)L_{\sigma}^2p_T
\max_i  E_{\mathbf{U}_{1}},\ldots,E_{\mathbf{U}_{T}}\\
[\|\acute{cv}_{\{T,\acute{T},1\}}(\sigma(\phi_{\{T,\acute{T}-1\}}(\mathbf{x}^i)))\|_2^2 ]\bigg)\bigg]^\frac{1}{2} \nonumber
\end{multline} 
Again using the technique as in eq. \ref{eq: VCRES1} we get,
\begin{multline}
= A_{P+1}\bigg[\left(L_\sigma^{2P}\prod_{k=1}^{P} p_kh_k A_{k}^2\right)\bigg(\left(A_{T}N_{T}v_T^2\right)\left(A_TN_Tv_T^2\right)^2L_{\sigma}^2p_T 
\max_i  E_{\mathbf{U}_{1}},\ldots,E_{\mathbf{U}_{T}}\\
[\|\acute{\sigma}(\phi_{\{T,\acute{T}-1\}}(\mathbf{x}^i))\|_2^2 ]\bigg)\bigg]^\frac{1}{2} \nonumber
\end{multline} 
\begin{multline}
= A_{P+1}\bigg[\left(L_\sigma^{2P}\prod_{k=1}^{P} p_kh_k A_{k}^2\right)\bigg(\left(A_{T}N_{T}v_T^2\right)\left(A_TN_Tv_T^2\right)^2L_{\sigma}^4p_T 
\max_i  E_{\mathbf{U}_{1}},\ldots,E_{\mathbf{U}_{T}}\\
[\|\acute{\phi}_{\{T,\acute{T}-1\}}(\mathbf{x}^i)\|_2^2 ]\bigg)\bigg]^\frac{1}{2} \nonumber
\end{multline} 
Doing the above for $\acute{T}$ residual units in each of the $T$ residual blocks we arrive at,
\begin{multline}
= A_{P+1}\bigg[\left(L_\sigma^{2P}\prod_{k=1}^{P} p_kh_k A_{k}^2\right)\bigg(\left(A_{0}N_{0}v_0^2\right)\prod_{r=1}^{T}\left(A_{r}N_{r}v_r^2\right)\left(A_rN_rv_r^2\right)^{2\acute{T}}L_{\sigma}^{4\acute{T}}p_r^{\acute{T}} \\
\max_i  [\|(\mathbf{x}^i)\|_2^2 ]\bigg)\bigg]^\frac{1}{2} \nonumber
\end{multline} 
Let $R$ denote the radius of the dataset, thus we get the following bound on VC dimension of a residual network:
\begin{multline}
VCdim(\mathcal{F})\leq R^2A_{P+1}^2\bigg[\left(L_\sigma^{2P}\prod_{k=1}^{P} p_kh_k A_{k}^2\right)\bigg(\left(A_{0}N_{0}v_0^2\right)\prod_{r=1}^{T}\left(A_rN_rv_r^2\right)^{3\acute{T}}L_{\sigma}^{4\acute{T}}p_r^{\acute{T}} \bigg)\bigg]  \nonumber
\end{multline}


\subsection*{Proof of Theorem \ref{th:5}} \label{pr:5}
\textbf{Proof}: Following the eq. \ref{eq: error1} - eq. \ref{eq: VC FNN3b} and replacing $x^i$ with $\mathbf{\hat{x}}^i = \mathbf{x}^i + \Delta^i$ we obtain,
\begin{gather}
\sqrt{m} \leq A_{P+1} \big[\max_i(\|\mathbf{\hat{x}}^i\|_2^2) L_\sigma^{2P}\prod_{k=1}^{P} h_k A_{k}^2 \big]^{\frac{1}{2}}\nonumber \\
         = A_{P+1} \big[\max_i(\|\mathbf{x}^i + \Delta^i\|_2^2) L_\sigma^{2P}\prod_{k=1}^{P} h_k A_{k}^2 \big]^{\frac{1}{2}} \label{eq: VC FNN9}
\end{gather}
Applying triangle inequality to the term $\|x^i + \Delta^i\|_2^2$ in eq. \ref{eq: VC FNN9} we get,
\begin{gather}
\sqrt{m} \leq A_{P+1} \big[\max_i(\|\mathbf{x}^i\|_2 + \|\Delta^i\|_2^2) L_\sigma^{2P}\prod_{k=1}^{P} h_k A_{k}^2 \big]^{\frac{1}{2}} \label{eq: VC FNN10}
\end{gather}
Since $\max_i(\|\mathbf{x}^i\|_2) \leq R^2$ and $\max_i(\|\Delta^i\|_2^2) \leq c^2$, using these in eq. \ref{eq: VC FNN10} we obtain,
\begin{gather}
\sqrt{m} \leq A_{P+1} \big[(R^2+c^2) L_\sigma^{2P}\prod_{k=1}^{P} h_k A_{k}^2 \big]^{\frac{1}{2}} \nonumber\label{eq: VC FNN11}\\
m \leq A_{P+1}^2 \big[(R^2+c^2) L_\sigma^{2P}\prod_{k=1}^{P} h_k A_{k}^2 \big] \nonumber \label{eq: VC FNN12}\\
\implies VCdim(\mathcal{F}_{ro}) \leq A_{P+1}^2 \big[(R^2+c^2) L_\sigma^{2P}\prod_{k=1}^{P} h_k A_{k}^2 \big]\nonumber \label{eq: VC FNN13}
\end{gather}
\vskip 0.2in
\bibliography{bibliography}

\end{document}